\documentclass[journal]{IEEEtran}

\usepackage{mathtools}
\usepackage{amsmath,amssymb,graphicx,booktabs}
\usepackage{subcaption}

\newcommand{\vect}[1]{\boldsymbol{\mathbf{#1}}}
\usepackage{color,soul}
\DeclareMathOperator{\E}{\mathbb{E}}

\ifCLASSINFOpdf
\else
\fi
\begin{document}
%
\title{An online sequence-to-sequence model for noisy speech recognition}
%
%
%

\author{
Chung-Cheng Chiu*,
Dieterich Lawson*\thanks{*Equal Contribution},
Yuping Luo,
George Tucker,
Kevin Swersky,
Ilya Sutskever,
Navdeep Jaitly}


%
%

\markboth{An online sequence-to-sequence model for noisy speech recognition}
{Chiu \MakeLowercase{\textit{et al.}}: online sequence-to-sequence model for noisy speech recognition}
%

\maketitle

\begin{abstract}
Generative models have long been the dominant approach for speech recognition.
The success of these models however relies on the use of sophisticated recipes
and complicated machinery that is not easily accessible to non-practitioners.
Recent innovations in Deep Learning have given rise to an alternative --
discriminative models called Sequence-to-Sequence models, that can almost match
the accuracy of state of the art generative models.
While these models are easy to train as they can be trained end-to-end in
a single step, they have a practical limitation that they can only be
used for offline recognition. This is because the models require that the entirety of the
input sequence be available at the beginning of inference, an assumption that is not valid
for instantaneous speech recognition.  To address this problem,
online sequence-to-sequence models were recently introduced.
These models are able to start producing outputs as data arrives, and the
model feels confident enough to output partial transcripts. These models,
like sequence-to-sequence are causal -- the output produced by the model
until any time, $t$, affects the features that are computed subsequently.
This makes the model inherently more powerful than generative models
that are unable to change features that are computed from the data.
This paper highlights two main contributions -- an improvement to
online sequence-to-sequence model training, and its application to
noisy settings with mixed speech from two speakers.
\end{abstract}

\begin{IEEEkeywords}
Automatic Speech Recognition, End-to-End Speech Recognition, Very Deep Convolutional Neural Networks
\end{IEEEkeywords}

%
\IEEEpeerreviewmaketitle

\section{Introduction}
\IEEEPARstart{G}enerative models have long been the bread and butter of traditional speech
recognition techniques. Using these models, transcription is typically performed
by Maximum-a-posteriori (MAP) estimation of the word sequence, given a trained
generative model and an acoustic observation. Gaussian Mixture Models (GMMs)
were the dominant models for instantaneous emission distributions and were
coupled to Hidden Markov Models (HMMs) to model the dynamics. While posteriors
from the GMM's have been supplanted by Deep Neural Networks (DNN) lately,
the recognition model essentially retains its generative interpretation.

Recent developments in deep learning have given rise to a powerful
alternative -- discriminative models called sequence-to-sequence
models, can be trained to model the conditional probability
distribution of the output transcript sequence given the input acoustic
sequence, directly without inverting a generative model.
Sequence-to-sequence models \cite{sutskever-nips-2014,cho-emnlp-2014}
are a general model family for solving supervised learning problems
where both the inputs and the outputs are sequences.  The performance
of the original sequence-to-sequence model has been greatly improved
by the invention of \emph{soft attention} \cite{bahdanau-iclr-2015},
which made it possible for sequence-to-sequence models to generalize
better and achieve excellent results using much smaller networks on
long sequences.  The sequence-to-sequence model with attention had
considerable empirical success on machine translation
\cite{bahdanau-iclr-2015}, speech recognition
\cite{chorowski-nips-2014,chan2015listen}, image caption generation
\cite{xu-icml-2015,vinyals-arvix-2014}, and question answering
\cite{weston2014memory}.

Although remarkably successful, the sequence-to-sequence model with
attention must process the entire input sequence before producing an
output.  However, there are tasks where it is useful to start
producing outputs before the entire input is processed. These tasks
include speech recognition, machine translation and simultaneous
speech recognition and translation with one model~\cite{babelfish}.

Recently new models have been developed that overcome these shortcomings.
These models, which we call {\it online sequence-to-sequence models}
have the property that they produce outputs as inputs are received
~\cite{jaitly2015online,luo2016learning}, while retaining the causal
nature of sequence-to-sequence models. In this paper, we use the
model that we previously introduced in~\cite{luo2016learning}\footnote{
We borrow text heavily from this prior paper to explain the motivation and
several details about the model}.
This model uses binary stochastic variables to select the timesteps at which
to produce outputs. We call this model the \emph{Neural Autoregressive
Transducer (NAT)}. The stochastic variables are trained with a policy
gradient method. However unlike the work by Luo et. al~\cite{luo2016learning}
we use a modified method of training that improves our training results.
Further, we explore the use of this model for noisy input where we
present single channel mixed speech from two speakers at different
mixing proportions as input to the model. This models is uniquely
suited for this task as it is a causal model, and as it is trained
discriminatively. We show results of this model on a task we call
MultiTIMIT it shows that the model is able to handle noisy speech
quite well. We speculate that the use of this model with a multiple
microphone arrangement should lead to strong results on mixed and
noisy speech.

\begin{figure*}[t]
  \centering
  \includegraphics[width=0.8\textwidth]{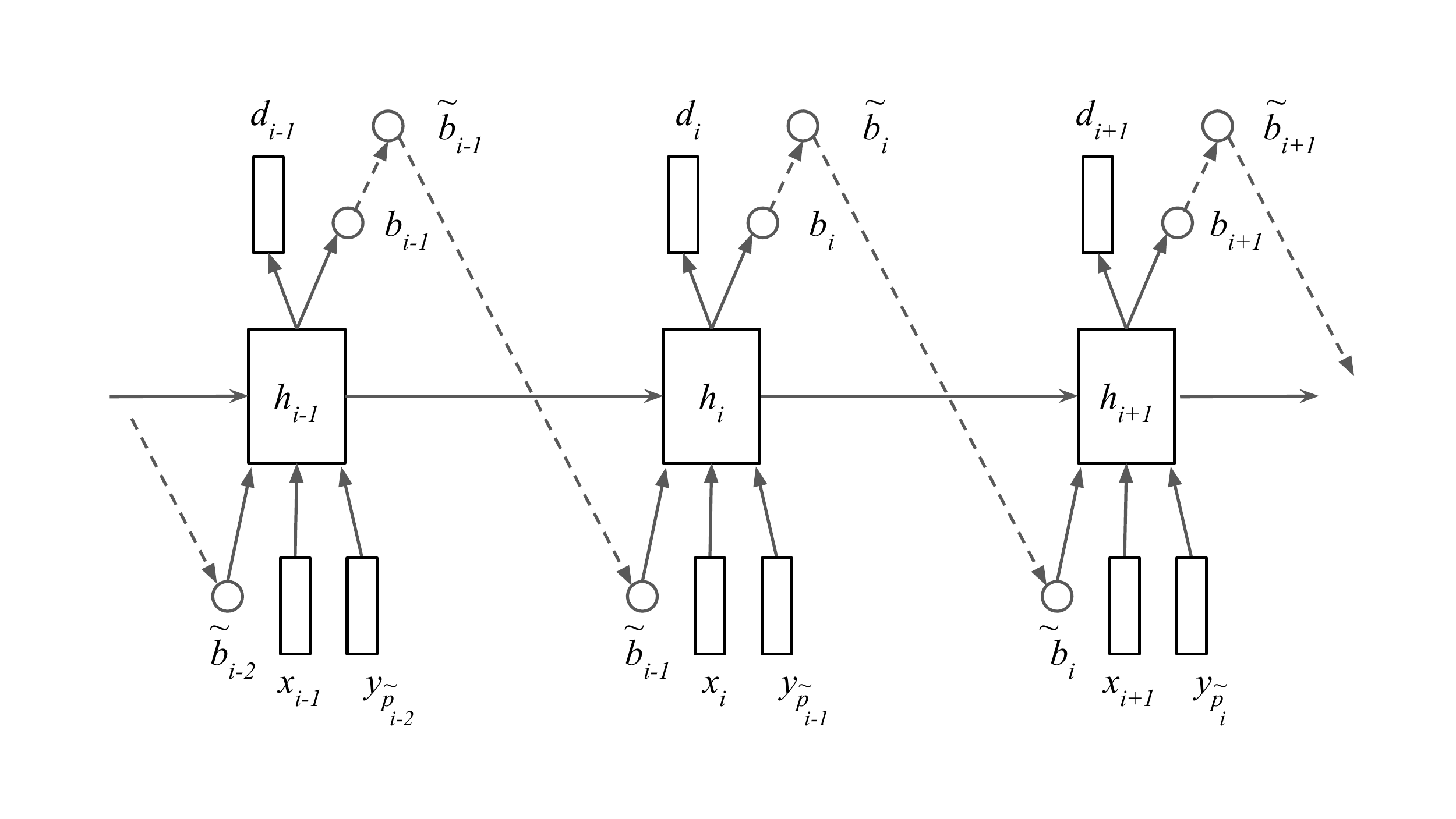}
  \caption{Overall Architecture of the model used in this paper.}\label{fig:model}
\end{figure*}
\subsection{Relation To Prior Work}

Sequence to sequence models have been recently applied to phoneme
recognition~\cite{chorowski-nips-2015} and speech recognition~\cite{chan2015listen}.
In these models, the input acoustics, in the form of log Mel filter banks
are processed with an encoder neural network that is usually a
bidirectional neural network. A decoder then produces output tokens
one symbol at a time, using next step prediction. At each step, the
decoder uses ``soft attention'' over the encoder time steps to create
a ``context vector'' that is a summary of features of the encoder.
The context vector is fed into the decoder and is used to make the
prediction at any time step.

While the idea of soft attention as it is currently understood was
first introduced by Graves \cite{graves2013generating}, the first
truly successful formulation of soft attention is due to Bahdanau et
al.~\cite{bahdanau-iclr-2015}. It used a neural architecture that
implements a ``search query'' that finds the most relevant element in
the input, which it then picks out.  Soft attention has quickly become
the method of choice in various settings because it is easy to implement
and it has led to state of the art results on various tasks. For example,
the Neural Turing Machine \cite{graves2014neural} and the Memory Network
\cite{sukhbaatar2015end} both use an attention mechanism similar to
that of Bahdanau et al.~\cite{bahdanau-iclr-2015} to implement models
for learning algorithms and for question answering.

While soft attention is immensely flexible and easy to use, it assumes
that the test sequence is provided in its entirety at test time.  It
is an inconvenient assumption whenever we wish to produce the relevant
output as soon as possible, without processing the input sequence in
its entirety first. Doing so is useful in the context of a speech
recognition system that runs on a smartphone, and it is especially
useful in a combined speech recognition and a machine translation
system.

This model can be thought of extending two previous models -- the Connectionist
Temporal Classification (CTC)~\cite{graves-icml-2014}
and the Sequence Transducer~\cite{graves-icml-2012} models that have
been used for speech recognition previously. However, neither CTC nor the
Sequence Transducer are causal models -- both models compute
features from the data independently at each time step, and this feature
computation is unaffected by the tokens output previously. Note that
while the language model RNN in the sequence transducer computes
predictions causally, these do not impact the local class predictions made by the
acoustics which are independent of each others and not causal.

There exists prior work that investigated causal models for producing an
output without consuming the input in its entirety.  These include the
work by Mnih \cite{mnih2014recurrent} and Zaremba and Sutskever
\cite{zaremba2015reinforcement} who used the Reinforce algorithm to
learn the location in which to consume the input and when to emit an
output.  Finally, Jaitly et al.~\cite{jaitly2015online} used an
online sequence-to-sequence method with conditioning on partial
inputs,  which yielded encouraging results on the TIMIT dataset.

This work is technically an extension of our prior work in~\cite{luo2016learning}
where policy gradients with continuous rewards was used to train
the model. In this paper, we use similar ideas, but instead of using
a single sample REINFORCE model with a parameteric baseline for
centering the training of the stochastic model, we use a multi-sample
training, with a baseline that is an average over leave-one-out samples.

Further, in this paper, we explore the use of this model for
noisy data -- specifically noisy data that corresponds to speech from
two different speakers mixed in at different levels.

\section{Methods}
In this section we describe the details of the Autoregressive
Sequence Transducer. This includes the recurrent neural
network architecture, the reward function, and the training and
inference procedure. Much of the description is borrowed heavily
from our description in~\cite{luo2016learning}. We refer the
reader to figure~\ref{fig:model} for the details of the model.

\begin{figure*}[t]
  \centering
  \includegraphics[width=\textwidth]{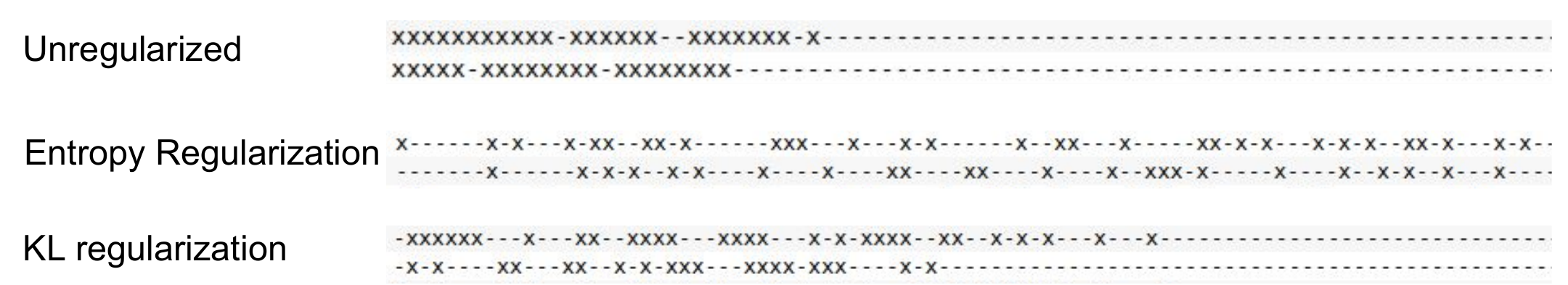}
  \caption{The impact of entropy regularization on emission locations. Each
     line shows the emission predictions made for an example input utterance,
     with each symbol representing 3 input time steps. 'x' indicates that the
     model chooses to emit output at the time steps, whereas '-' indicates
     otherwise. Top line - without entropy penalty the model emits symbols
     either at the start or at the end of the input, and is unable to get
     meaningful gradients to learn a model. Middle line - with entropy
     regularization, the model avoids clustering emission predictions in
     time and learns to spread the emissions meaningfully and learn a model.
     Bottom line - using KL divergence regularization of emission probability
     also mitigates the clustering problem, albeit not as effectively as with
     entropy regularization.}\label{fig:entropy_pen}
\end{figure*}

We begin by describing the probabilistic model we used in this
work. 
At each time step, $i$, a recurrent neural network (represented in figure 1) decides whether to emit an output token. The decision is made by a stochastic binary logistic unit $b_i$. Let $\tilde{b}_i \sim \text{Bernoulli}(b_i)$ be a Bernoulli distribution such that if $\tilde{b}_i$ is 1, then the model outputs the vector $d_i$, a softmax distribution over the set of possible tokens. The current position in the output sequence $\mathbf{y}$ can be written $\tilde{p}_i = \sum_{j=1}^i \tilde{b}_j$, which is incremented by 1 every time the model chooses to emit. Then the model's goal is to predict the desired output $y_{\tilde{p}_i}$; thus whenever $\tilde{b}_i = 1$, the model experiences a loss given by
\[\text{softmax\_logprob}(d_i;y_{\tilde{p}_i}) = - \sum_{c}\log(d_{ic})y_{\tilde{p}_i c}\]
where $c$ ranges over the number of possible output tokens.


At each step of the RNN, the binary decision of the previous timestep,
$\tilde{b}_{i-1}$ and the corresponding previous target $t_{i-1} = y_{\tilde{p}_{i-1}}$
are fed into the model as input.  This feedback ensures that the model's
outputs are causally dependent on the model's previous outputs, and thus
the model is from the sequence to sequence family.

We train this model by estimating the gradient of the log probability
of the target sequence with respect to the parameters of the
model. While this model is not fully differentiable because it uses
non-diffentiable binary stochastic units, we can estimate the
gradients with respect to model parameters by using a policy gradient
method, which has been discussed in detail by Schulman et
al.~\cite{schulman2015gradient} and used by Zaremba and Sutskever
\cite{zaremba2015reinforcement}.

In more detail, we use supervised learning to train the network to
make the correct output predictions, and reinforcement learning to
train the network to decide on when to emit the various
outputs.  Let us assume that the input sequence is given by $(x_1,\ldots,x_{T_1})$
and let the desired sequence be $(y_1,\ldots,y_{T_2})$, where $y_{T_2}$ is a special
end-of-sequence token, and where we assume that $T_2 \leq T_1$. Then the log
probability of the model is given by the following equations:
\begin{eqnarray}
  h_i &=& \textrm{LSTM}(h_{i-1}, \mathrm{concat}(x_i, \tilde{b}_{i-1}, \tilde{y}_{i-1})) \\
  b_i &=& \mathrm{sigmoid}(W_\mathrm{b} \cdot h_i) \\
  \tilde{b}_i &\sim & \mathrm{Bernoulli}(b_i) \\
  \tilde{p}_i &=& \sum_{j=1}^i \tilde{b}_j \\
  \tilde{y}_i &=& y_{\tilde{p}_i} \\
	d_i &=& \mathrm{softmax}(W_o h_i) \\
  \mathcal{R}  &=& \mathcal{R} + \tilde{b}_i \cdot \text{softmax\_logprob}(d_i; \tilde{y}_i)  \label{eqn:reward}
\end{eqnarray}
\begin{figure*}[t!]
  \centering
    \begin{subfigure}{.3\linewidth}
        \centering
        \includegraphics[width=\linewidth]{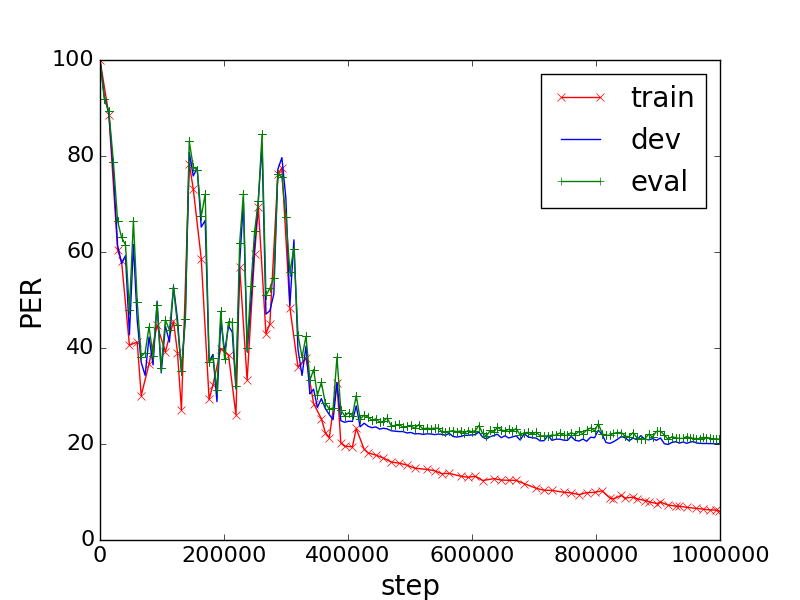}
        \caption{Clean TIMIT}\label{fig:timit}
    \end{subfigure}%
    \begin{subfigure}{.3\linewidth}
        \centering
        \includegraphics[width=\linewidth]{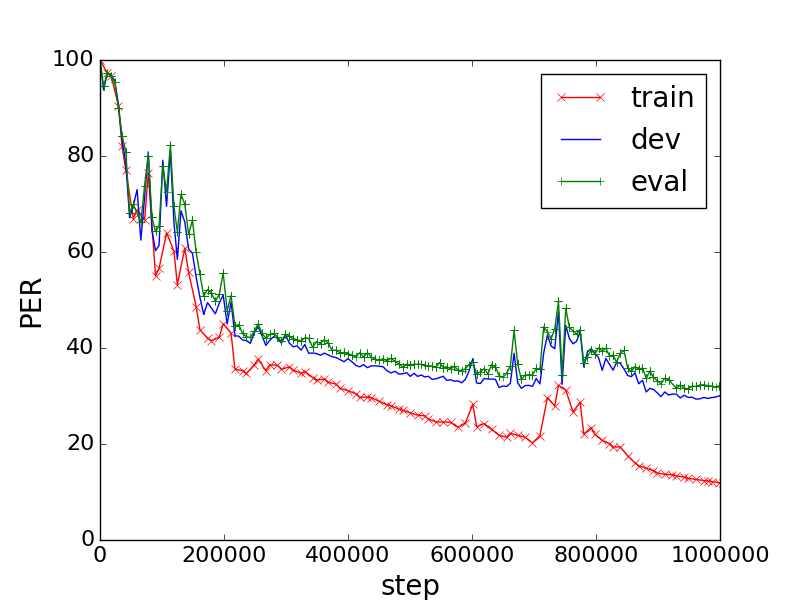}
        \caption{Multi-speaker - with 25\% mixing}\label{fig:timit_25}
    \end{subfigure}%
    \begin{subfigure}{.3\linewidth}
        \centering
        \includegraphics[width=\linewidth]{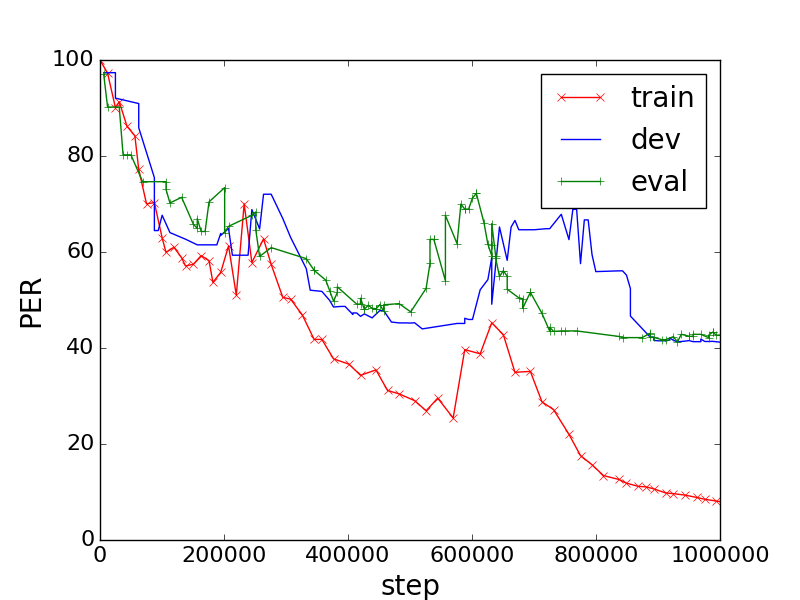}
        \caption{Multi-speaker - with 50\% mixing}\label{fig:timit_50}
    \end{subfigure}
  \caption{Example training run on TIMIT.}\label{fig:timit}
\end{figure*}

In the above equations, $\tilde{p}_i$ is the ``position'' of the model
in the output, which is always equal to $\sum_{k=1}^i \tilde{b}_i$:
the position advances if and only if the model makes a prediction.
Note that we define $y_0$ to be a special beginning-of-sequence
symbol.  The above equations also suggest that our model can easily be implemented
within a static graph in a neural net library such as TensorFlow~\cite{abadi2016tensorflow}, even though
the model has, conceptually, a dynamic neural network architecture.

Following Zaremba and Sutskever~\cite{zaremba2015reinforcement}, we modify the model from
the above equations by forcing $\tilde{b}_i$ to be equal to 1
whenever $T_1 - i \leq T_2 -\tilde{p}_i$.  Doing so ensures that
the model will be forced to predict the entire target sequence $(y_1,\ldots,y_{T_2})$,
and that it will not be able to learn the degenerate solution where
it chooses to never make any prediction and therefore never experience
any prediction error.

We now elaborate on the manner in which the gradient is computed.  It
is clear that for a given value of the binary decisions $\tilde{b}_i$,
we can compute $\partial \mathcal{R} /\partial \theta$ using the
backpropagation algorithm.  Figuring out how to learn $\tilde{b}_i$ is
slightly more challenging.  To understand it, we will factor the
reward $\mathcal{R}$ into an expression $\mathcal
{R}(\tilde{\vect{b}})$ and a distribution $\rho(\tilde{\vect{b}})$
over the binary vectors, and derive a gradient estimate with respect
to the parameters of the model:
\begin{equation}
\label{eqn:expectedR}
\mathcal{R} = \E_{\vect{\tilde{b}}} \left[R(\tilde{\vect{b}}) \right]
\end{equation}

Differentiating, we get
\begin{equation}
\label{eqn:grad_reward}
\nabla \mathcal{R} = \E_{\vect{\tilde{b}}} \left[ \nabla R(\tilde{\vect{b}})
       + R(\tilde{\vect{b}}) \nabla \log \rho(\tilde{\vect{b}})
 \right]
\end{equation}
where $\rho(\tilde{\vect{b}})$ is the probability of a binary sequence of the $\tilde{b}_i$
decision variables. In our
model, $\rho(\tilde{\vect{b}})$ is computed using the chain rule over the $b_i$ probabilities:
\begin{equation}
\label{eqn:emit_prob}
\log \rho (\tilde{\vect{b}}) = \sum_{i=1}^T \tilde b_i \log b_i + (1-\tilde b_i) \log (1-b_i)
\end{equation}

Since the gradient in equation~\ref{eqn:grad_reward} is a policy
gradient, it has very high variance, and variance reduction techniques
must be applied. As is common in such problems we use {\it centering}
(also known as baselines) and Rao-Blackwellization to reduce the
variance of such models. See Mnih and Gregor \cite{anvil} for an example of the use of
such techniques in training generative models with stochastic units.

Baselines are commonly used in the reinforcement learning literature to 
reduce the variance of estimators, by relying on the identity
$\E_{\vect{\tilde{b}}}  \left[ \nabla \log \rho(\tilde{\vect{b}}) \right] = 0$. Thus
the gradient in~\ref{eqn:grad_reward} can be better estimated by the following, through the
use of a well chosen {\it baseline} function, $\Omega({\bf x})$, where ${\bf x}$ is a vector
of side information which happens to be the input and all the outputs up to timestep $\tilde{p}_i$:
\begin{equation}
\label{eqn:baselined_reward}
\nabla \mathcal{R} = \E_{\vect{\tilde{b}}} \left[ \nabla R(\tilde{\vect{b}})
       + \left(R(\tilde{\vect{b}}) - \Omega(\bf{x})\right) \nabla \log \rho (\tilde{\vect{b}})
 \right]
\end{equation}
The variance of this estimator itself can be further reduced by Rao-Blackwellization,
giving:
\begin{align}
\E_{\vect{\tilde{b}}} & \left[\left(R(\tilde{\vect{b}}) - \Omega(\bf{x})\right)
\nabla \log \rho (\tilde{\vect{b}}) \right] = \nonumber \\
 &\sum_{j=1}^T \E_{\vect{\tilde{b}}} \left[ \left( \sum_{i=j}^T R_i - \Omega_j \right)
       \nabla \log p (b_j | b_{<j}, \vect{x}_{\leq j}, \vect{y}_{\leq \tilde{p}_{j-1}}) \right]
\end{align}

This above term, while not computable analytically, can be estimated numerically
by drawing $K$ sample trajectories, indexed by $k$. Thus we have an estimate
of the gradient as follows:
\begin{multline}
\nabla \mathcal{R} \approx \\ \frac{1}{K} \sum_{k=1}^K \sum_{j=1}^T \left[\left( \sum_{i=j}^T R_i^{k} - \Omega_j^k \right)
       \nabla \log p (b_j^k | b_{<j}^k, \vect{x}_{\leq j}, \vect{y}_{\leq \tilde{p}_{j-1}}) \right]
\end{multline}
where, the superscript of $k$ indicates the sample index.
In previous work~\cite{luo2016learning} we used a single sample estimate (i.e. $K=1$)
and a neural network as a parametric baseline to estimate $\Omega_j^k$. This was
computed using a linear projection of the hidden state $h_j$ of the top LSTM layer of
the RNN, i.e. $\Omega_j^k = W' h_j^k + o$, where $W$ is a vector and $o$ is a bias.

Recent work in reinforcement learning and variational methods has shown the advantage
of multi-sample estimates~\cite{burda2015importance,vimco}.
In this paper, we thus explore the use of a multi-sample estimate, with $K=16$. Further,
as in~\cite{vimco} we used a baseline with a leave one out average, which we explain
next.

A straightforward choice of this baseline, $\Omega_j^k$ is the average sum of future
rewards from the other samples
\begin{equation*}
    \Omega_j^k = \frac{1}{K-1} \sum_{k' \neq k} \sum_{i \geq j}^T R^{k'}_{i},
\end{equation*}
however, this ignores the fact that the internal state of the different samples
are not the same. Ideally we would average over multiple trajectories starting
from the same state (i.e. number of inputs consumed and outputs produced), but
this is computationally expensive. As a result there is an imbalance where some
of the samples have emitted more symbols than the others, and thus the future
rewards may not be directly comparable. We add a residual term to address this,
\begin{equation}
\Omega_j^k = \frac{1}{K-1} \sum_{k' \neq k} \sum_{i \geq j}^T R^{k'}_{i} + \frac{1}{K-1} \sum_{k' \neq k} \sum_{i < j}  (R^{k'}_{i} - R^{k}_{i}) 
\label{eq:vimco_c}
\end{equation}
We call this the \emph{leave-one-out} baseline.

Finally, we note that reinforcement learning models are often trained with augmented
objectives that add an entropy penalty for actions are the too
confident~\cite{levine2014motor,williams1992simple}. We found this to be crucial for our models to train
successfully. In light of the regularization term, the augmented reward at any time
steps, $i$, is:
\begin{multline}
\label{eqn:entropy_regularization}
R_i = \tilde{b}_i \log p(d_i = t_i | \vect{x}_{\leq i},\tilde{\vect{b}}_{<i},
\vect{t}_{<i}) \\ - \lambda \tilde{b}_i \log p (b_i=1 | b_{<i}, \vect{x}_{\leq i})  \\
              + \lambda (1-\tilde{b}_i) \log (p (b_i=0 | b_{<i}, \vect{x}_{\leq i}))
\end{multline}

Without the use of this regularization in the model, the RNN emits all the symbols
clustered in time, either at very start of the input sequence, or at the end.
The model has a difficult time recovering from this configuration, since the gradients
are too noisy and biased. However, with the use of this penalty, the model successfully
navigates away from parameters that lead to very clustered predictions and eventually
learns sensible parameters. An alternative we explored was to use the the KL divergence
of the predictions from a target Bernouilli rate of emission at every step. However,
while this helped the model, it was not as successful as entropy regularization. See
figure~\ref{fig:entropy_pen} for an example of this clustering problem and how
regularization ameliorates it.

\section{Experiments and Results}
\begin{figure*}[th]
\vskip 0.2in
\begin{center}
\includegraphics[width=\linewidth]{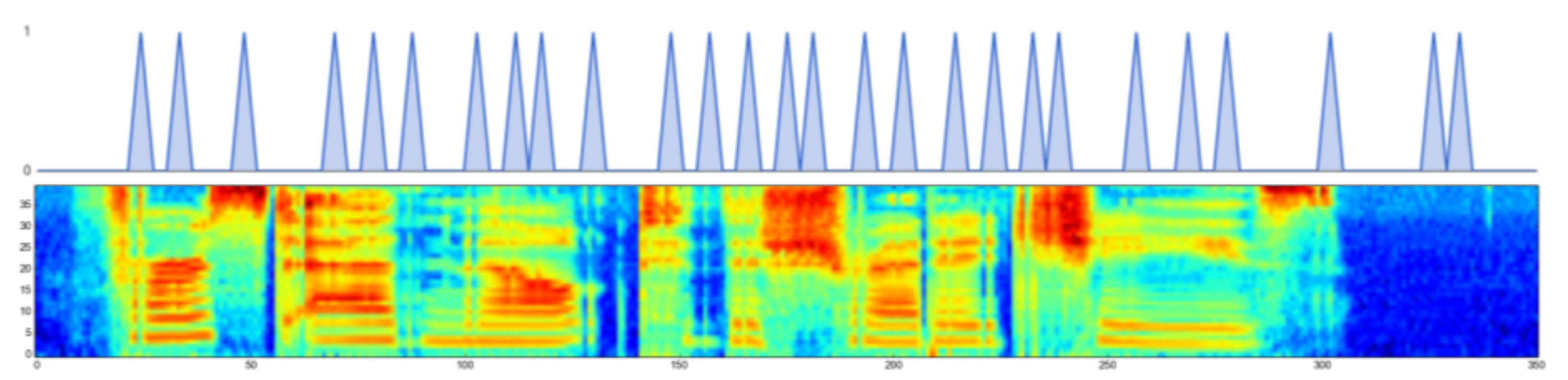}
\caption{This figure shows the model emission distributions, the probability of emitting tokens as the input is received. The target phonemes are ``pau f er s pau t pau ae pau m ih l pau t ah dh ih sh r eh dx ih pau ch iy s pau''}
\label{emissions}
\end{center}
\vskip -0.2in
\end{figure*} 
We conducted experiments on two different speech corpora using this model. Initial experiments were
conducted on TIMIT to assess hyperparameters that could lead to stable behavior of the model. The
second set of experiments were conducted on speech mixed in from two different speakers -- a male
speaker and a female speaker -- at different mixing proportions. We call these experiments 
Multi-TIMIT.

\subsection{TIMIT}\label{sec:exp_timit}
The TIMIT data set is a phoneme recognition task in which phoneme sequences have to be inferred from
input audio utterances. The training dataset contains 3696 different audio clips and the target is
one of 60 phonemes. Before scoring, these are collapsed to a standard 39 phoneme set, and then the
Levenshtein edit distance is computed to get the phoneme error rate (PER).

The models we trained on TIMIT had two layers with 256 units per layer. Each model was trained with
Adam(~\cite{kingma2014adam}) and used a learning rate of $7 \times 10^{-5}$. We used asynchronous SGD with
16 replicas in tensorflow as the neural network framework for training the models~\cite{abadi2016tensorflow,dean-nips-2012}.
No GPUs were used in the training.

Entropy regularization was crucial to produce the best results, with emissions clumping at the end
or beginning of the utterances when an entropy penalty was not used. We started the weight of the
entropy penalty at 1 and decayed it linearly to 0.1. We began decaying the entropy penalty at
10,000 steps and experimented with ending the decay at \{100,000, 200,000, 300,000, 400,000\} steps,
finding that step 200,000 worked best. After step 200,000 the entropy penalty weight was kept at 0.1.

We also regularized our models with variational weight noise~\cite{graves2011practical}. We tested
the values \{0.075, 0.1, 0.15\} for the standard deviation of the noise and found that 0.15 worked
best. We started the standard deviation of the variational noise at 0, and increased it linearly
from step 10,000 to a value of 0.15 at step 200,000. In each experiment the entropy penalty
stopped decaying on the same step that the variational noise finished increasing.

We also used L2-norm weight regularization to encourage small weights. We found that a weight of 0.001
worked best after trying weights $\left\{10^{-5}, 10^{-4}, 10^{-3}\right\}$.



Lastly, we note that the input filterbanks were processed such that three continuous frames of
filterbanks, representing a total of 30ms of speech were concatenated and input to the model.
This results in a smaller number of input steps and allows the model to learn hard alignments
much faster than it would otherwise.

See Figure~\ref{fig:timit} for an example of a training curve. It can be seen that the model
requires a larger number of updates ($\textgreater$ 100K) before meaningful models are learnt. However, once
learning starts, steady process is achieved, even though the model is trained by policy gradient.

Table~\ref{tab:timit} shows a summary of the results achieved on TIMIT by our method and
other, more mature models. As can be seen our model compares favorably with other unidirectional
models, such as CTC, DNN-HMM's etc. Combining with more sophisticated features such as convolutional
models should produce better results. Moreover, this model has the capacity to absorb language
models, and as a result, should be more suited to end to end training than CTC and DNN-HMM based
models that cannot inherently capture language models because they predict all tokens independently
of each other.
\begin{table}[h]
  \caption{Results on TIMIT using Unidirectional LSTMs for various models.}
  \label{tab:timit}
  \centering
  \begin{tabular}{lll}
    \toprule
    Method     & PER   \\
    \midrule
    CTC\cite{graves2013speech}     & 19.6\% \\
    DNN-HMM\cite{mohamed2012acoustic}     & 20.7\% \\
    seq2seq with attention (our implementation)     & 24.5\% \\
    neural transducer\cite{jaitly2015online}     & 19.8\% \\
    \midrule
    NAT (Stacked LSTM) + Parameteric Baseline\cite{luo2016learning}     & 21.5\% \\
    NAT (Grid LSTM) + Parameteric Baseline\cite{luo2016learning}     & 20.5\% \\
    NAT (Stacked LSTM) + Averaging Baseline (this paper) & 20.0\% \\
    \bottomrule
  \end{tabular}
\end{table}

\begin{figure*}[t]
\vskip 0.2in
\begin{center}
\includegraphics[width=\linewidth]{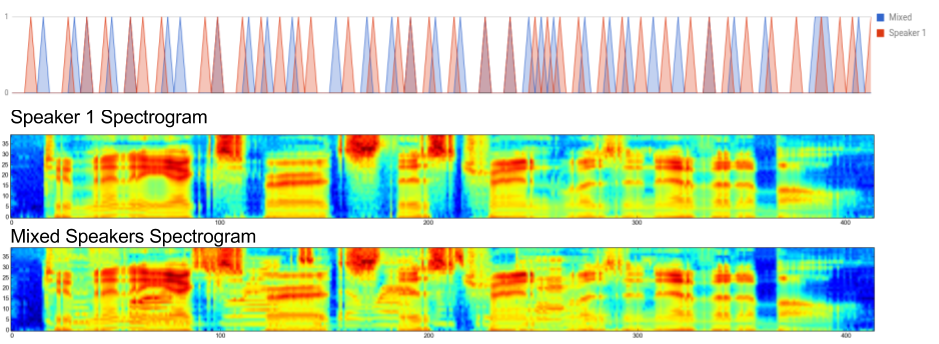}
\caption{Emission distributions for Multi-TIMIT: This figure shows the probability
of emitting tokens for the case of a clean utterance from TIMIT and a corresponding
noisy utterance in Multi-TIMIT. It can be seen that for the Multi-TIMIT utterances,
the model chooses to emit symbols slightly later than it would have for TIMIT utterances.}
\label{fig:mtimit_emission}
\end{center}
\vskip -0.2in
\end{figure*} 

\subsection{Multi-TIMIT}
We generate a new data set by mixing a male voice with a female voice from the original TIMIT data.  Each utterance in the original TIMIT data pairs with an utterance coming from the opposite gender.  The wave signal of both utterances are first scaled to the same range, and then the signal scale of the second utterance is reduced to a smaller volume when mixing the two utterances.  We explored different scale for mixing the second utterance, 50\%, 25\%, and 10\%, and created three sets of experiments.  The same feature generation method that was described above was used, resulting in a 123 dimensional input per frame. The transcript of the speaker 1 was used as the ground truth transcript for this new utterance. This data follow the same train, dev, and test specification as TIMIT. As a result the mixed data has the same number of train, dev, and test utterances as the original TIMIT, and they also have the same sets of target phonemes.

Our model was a 2-layer LSTM with 256 units in each layer. The same hyper-parameter search
strategy that was used for clean TIMIT (section~\ref{sec:exp_timit}) was applied here.

\begin{table}[h]
  \caption{Results on Multi-TIMIT: This table show the phoneme error rate (PER)
   achieved by our models at different proportions of mixing in for the distracting
   speech. Also shown are results from CTC with deep LSTMs~\cite{graves-icml-2014}
   and RNN-Transducer~\cite{graves-icml-2012} using an implementation provided by
   Alex Graves.}
  \label{tab:multitimit}
  \centering
  \begin{tabular}{cccc}
    \toprule
    Mixing Proportion     & NAT & CTC & RNN-Transducer   \\
    \midrule
    0.1  & 25.9\%  & 27.3\% & 25.7\% \\
    0.25 & 32.5\%  & 33.3\% & 32.2\% \\
    0.5  & 42.9\%  &43.8\% & 48.9\% \\
    \bottomrule
  \end{tabular}
\end{table}

Figures~\ref{fig:timit_25} and ~\ref{fig:timit_50} show examples of training curves
for two cases with mixing proportions of 0.25 and 0.5 respectively. In both cases it
can be seen that the model learns to overfit the data. 

Table~\ref{tab:multitimit} shows results from using different mixing proportions of confounding
speaker. It can be seen that with increasing mixing proportion, the model's results get worse
as expected. For the experiments, each audio input is always paired with the same confounding
audio input. Interestingly we found that pairing the same audio with multiple confounding
audio inputs produced worse results, because of much worse overfitting. This presumably
happens because our model is powerful enough to memorize the entire transcripts.

Figure~\ref{fig:mtimit_emission}  shows an example of where the model emits symbols for
an example Multi-TIMIT utterance. It also shows a comparison with the emissions from a
clean model. Generally speaking the model chooses to emit later for Multi-TIMIT compared
to when it emits for TIMIT.

\section{Discussion}
In this paper we have introduced a new way to train online sequence-to-sequence
models and showed its application to noisy input. These models, as a result
of being causal models, can incorporate language models, and can also generate
multiple different transcripts for the same audio input. This makes it
a very powerful class of models. Even on a dataset as small as TIMIT
the model is able to adapt to mixed speech. For our experiments each
speaker was only coupled to one distracting speaker and hence the dataset size
was limited. By pairing each speaker with multiple other speakers, and
predicting each one as outputs, we should be able to achieve greater
robustness. Because of this capability, we would like to apply these
models to multi-channel, multi-speaker recognition in the future.

\section{Conclusions}
In this work, we presented a new way of training an online sequence
to sequence model. This model allows us to exploit the modelling power
of sequence-to-sequence problems without the need to process the entire
input sequence first. We show the results of training this model on the
TIMIT corpus and acheived results comparable to state of the art
results with uni-directional models. 

We also applied this model to the task of mixed speech from two speakers,
producing the output for the louder speaker. We show that the model is
able to achieve reasonble accuracy, even with single channel input.
In the future, we will apply this work to multi-speaker recognition.

\ifCLASSOPTIONcaptionsoff
  \newpage
\fi



\bibliographystyle{IEEEtran}
\bibliography{IEEEabrv,paper}
%



%

\begin{IEEEbiographynophoto}{Chung-Cheng Chiu}
Chung-Cheng Chiu is a Software Engineer at Google Brain working on deep learning
models. He received his PhD from the University of Southern California under the
supervision of Stacy Marsella. His interests lie in Deep learning, speech recognition,
and reinforcement learning.
\end{IEEEbiographynophoto}%
\begin{IEEEbiographynophoto}{Dieterich Lawson}
Dieterich Lawson is a Brain Resident at Google working on sequential latent variable
models and methods for variational inference. He did his undergrad and masters at Stanford
in computer science and computational math, respectively. His interests include deep
learning, reinforcement learning, and optimization.
\end{IEEEbiographynophoto}%
\begin{IEEEbiographynophoto}{Yuping Luo}
Yuping Luo is a student at Tsinghua University majoring in Computer Science. His
interests are Deep Learning, Optimization, and related theory. He worked on Streaming
Algorithms and Online Learning at Tsinghua University under the supervision of
Periklis Papakonstantinou. He interned at Google Brain working on Speech Recognition
and was supervised by Ilya Sutskever and Navdeep Jaitly.
\end{IEEEbiographynophoto}%
\begin{IEEEbiographynophoto}{George Tucker}
George Tucker is a research software engineer at Google Brain working on deep learning models for sequences.
Prior to joining Google, he was a research scientist at Amazon working on deep acoustic models for
small-footprint keyword spotting. He received his PhD from MIT under the supervision of Bonnie Berger.
His interests lie in Deep Learning, Variational Inference, and Reinforcement Learning.
\end{IEEEbiographynophoto}%
\begin{IEEEbiographynophoto}{Kevin Swersky}
Kevin Swersky is a research scientist at Google brain. He received his PhD from the University
of Toronto under the supervision of Richard Zemel. His research interests include deep learning,
graphical models, generative models, and meta-learning. During his PhD, Kevin co-founded
Whetlab, an online hyperparameter tuning service, which was subsequently acquired by Twitter.
\end{IEEEbiographynophoto}%
\begin{IEEEbiographynophoto}{Ilya Sutskever}
Ilya Sutskever is the research director of OpenAI.  Previously, he was a research scientist at Google
Brain. He was also a co-founder of DNNResearch which was acquired by Google.  Sutskever has made
many contributions to the field of Deep Learning, including the first large scale convolutional
neural network that convincingly outperformed all previous vision systems by winning the 2012
ImageNet competition. He was listed in MIT Technology Review’s 35 innovators under 35.
\end{IEEEbiographynophoto}%
\begin{IEEEbiographynophoto}{Navdeep Jaitly}
Navdeep Jaitly is a Research Scientist at NVIDIA Research. Previously he was a research
scientist at Google Brain.  His interests lie in End-to-end models for speech recognition
and speech synthesis, Reinforcement Learning and new models for sequences using Deep Learning.
\end{IEEEbiographynophoto}




\end{document}